\documentclass[10pt,twocolumn,letterpaper]{article}

\usepackage[pagenumbers]{cvpr} % To force page numbers, e.g. for an arXiv version

\usepackage[dvipsnames,table]{xcolor}   % 'table' enables \rowcolor
\usepackage{booktabs}
\usepackage{graphicx}
\usepackage{tcolorbox}
\usepackage{xcolor}
\tcbset{probebox/.style={colback=skillbg!25, colframe=skillbg!80!black, boxrule=0.5pt, arc=1pt, left=3pt, right=3pt, top=2pt, bottom=2pt, fontupper=\footnotesize}}
\definecolor{skillbg}{HTML}{E6CDF6}

\newcommand{\gain}[1]{\textsubscript{\textcolor{ForestGreen}{+#1}}}
\newcommand{\loss}[1]{\textsubscript{\textcolor{BrickRed}{$-$#1}}}

\usepackage{microtype}

\definecolor{cvprblue}{rgb}{0.21,0.49,0.74}
\usepackage[pagebackref,breaklinks,colorlinks,allcolors=cvprblue]{hyperref}

\def\confName{CVPR}
\def\confYear{2027}

\title{\textsc{Skill.png}: Learning Agent Skills in Pixel Space}

\author{Maximilian Seeth\\
Hasso Plattner Institute\\
Technical University of Munich\\
{\tt\small max.seeth@tum.de}
\and
Chengzhi Martin Hu\\
Technical University of Munich\\
Helmholtz AI \\
{\tt\small cm.hu@tum.de}
\and
Marie Kempkes\\
Volkswagen Group Innovation\\
{\tt\small marie.kempkes@volkswagen.de}
\and
Paul Röttger\\
Hasso Plattner Institute\\
{\tt\small paul.rottger@hpi.de}
\and
Leo Schwinn\\
Technical University of Munich\\
Helmholtz AI \\
{\tt\small leo.schwinn@helmholtz-munich.de}
}

\begin{document}
\maketitle
\begin{abstract}
    \looseness=-1
    AI agents are commonly specialized for new tasks and domains using skills, i.e.\ natural language instructions placed in their context.
    However, the discrete nature of textual skills prevents continuous optimization.
    Gradient-based approaches such as finetuning, on the other hand, are typically model-specific, preventing skill transfer across different agents.
    Since many agents today are built on vision-language models (VLMs), we can address these limitations by introducing \textsc{Skill.png}:
    a method for obtaining skill-like image inputs whose pixels are optimized with gradient descent through the vision encoder of a frozen VLM.
    Image skills are compatible with any VLM, and a single image can be optimized jointly for an ensemble of models.
    We show that \textsc{Skill.png} significantly outperforms state-of-the-art textual skills optimized using frontier LLMs in 9 of 16 settings (across 4 VLMs and 4 benchmarks) while requiring only 64 to 256 image tokens instead of roughly 900 to 2000 text tokens at inference time.
    Hybrid skills, i.e.\ text and image skills jointly optimized, further improve agent performance.
    Moreover, an image optimized with an ensemble of VLMs can successfully transfer across the models in the ensemble.
    \textsc{Skill.png} thus presents the first agent skill that can be optimized with gradient descent while retaining transfer benefits of text skills. Our code is available at: \href{https://github.com/omseeth/skill.png}{https://github.com/omseeth/skill.png}
\end{abstract}    
\section{Introduction}
\label{sec:intro}

\begin{figure}[t]
    \centering
    \includegraphics[width=\columnwidth]
        {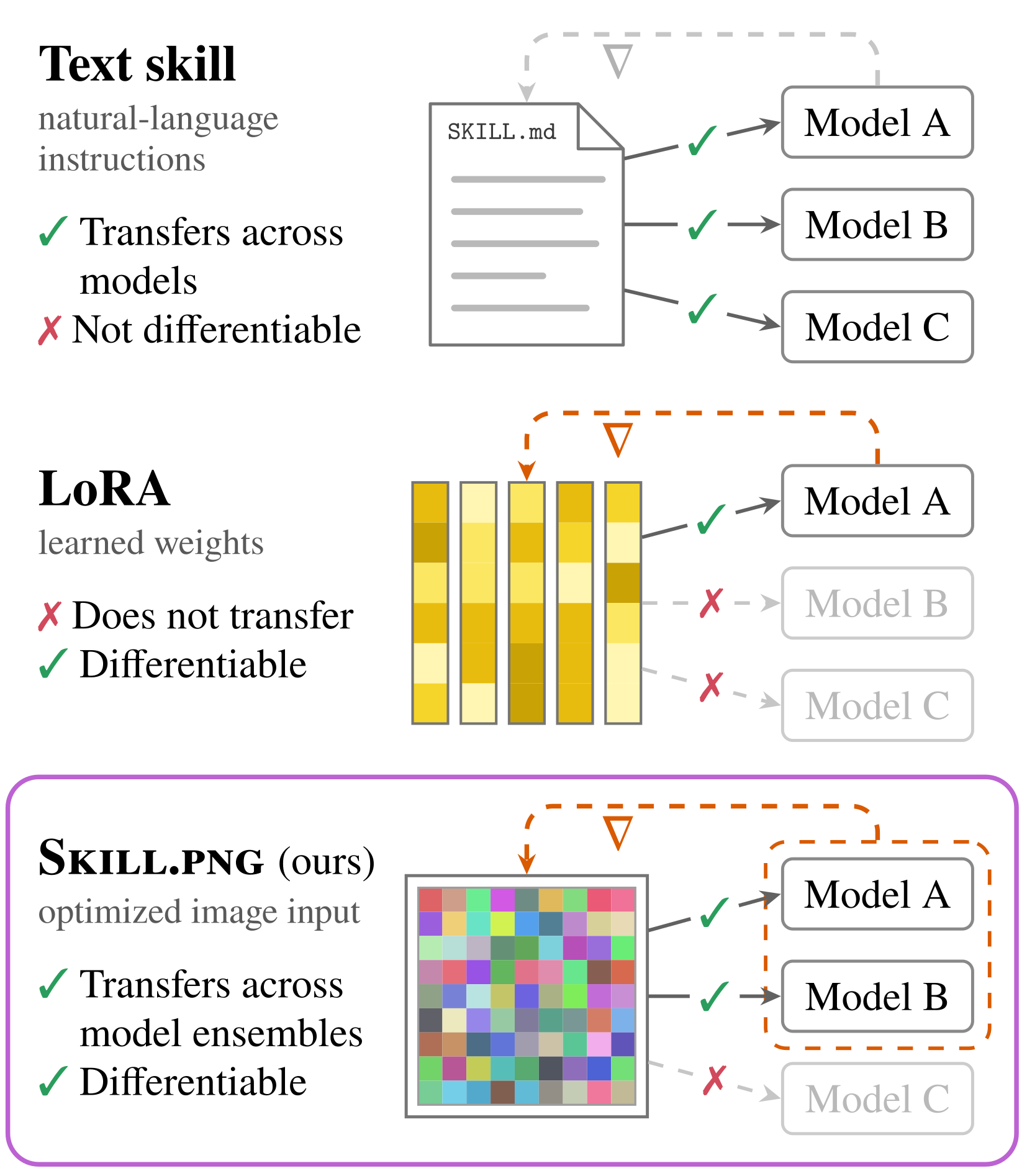}
    \caption{\textbf{Three ways to improve agents.}
    \emph{Text skills} work with any model but cannot be optimized with gradients. \emph{LoRA} is trained with gradients, but is model-specific. \textbf{\textsc{Skill.png} (ours)} encodes the skill as an image, optimized with gradients through one or more frozen models, and transfers across them.}
    \label{fig:overview}
\end{figure}

Skills, natural language instructions placed in a model's context, have become a common way to specialize agents for new tasks, tools and domains. Beyond writing them by hand, recent work refines skills in loops where a large language model (LLM) proposes edits and keeps those that help the agent \cite{agrawal2026gepa,alzubi2026evoskill,ni-etal-2026-trace2skill,yang2026skilloptexecutivestrategyselfevolving}. Relying on large and capable, often proprietary LLM optimizers is costly, though, compared to gradient-based optimizations, which are much more sample-efficient, as shown in adversarial research \cite{schwinn2024soft, xhonneux2024efficient, carlini}.

\looseness=-1
One way to bypass discrete search is to optimize in continuous parameter or embedding spaces instead. Finetuning with low-rank adaptation (LoRA) \cite{hu2022lora} learns updates of a model's weights, and visual prompt tuning \cite{Jia-Visual-Prompt-Tuning2022} learns continuous prompt tokens in a model's embedding space. Both methods are model-specific, either by merging new weights or inserting embeddings, requiring white-box access. However, parameters from a model's latent space are not skill artifacts that can be easily handed to other agents via its input.

Therefore, we introduce \textsc{Skill.png}, which brings gradient-based optimization to skills while staying in the input space. Many agents are built on vision-language models (VLMs), and every VLM accepts images. \textsc{Skill.png} continuously optimizes the pixels of an image through the vision encoder of a frozen VLM and keeps them in the valid image domain (Fig.~\ref{fig:overview}). A model provider can, for example, optimize skill images for its newest model in-house and release them alongside it. Users then simply attach these images, potentially along with text skills, to their queries without the need to access the agent's weights again. For open-weight models, skills can also be optimized and shared by users directly. 

Across four VLMs and four benchmarks spanning question answering, tool use, research-level mathematics and planning, \textsc{Skill.png} outperforms text skills from SkillOpt, an LLM optimization loop \cite{yang2026skilloptexecutivestrategyselfevolving}, in 9 of 16 settings significantly. It improves on the model without a skill by up to 37.8 benchmark points, while requiring only 64 to 256 image tokens instead of roughly 900 to 2000 text tokens. Text and image skills can also be combined: optimizing an image with a text skill already in context adds 6.1 to 15.5 points on three of four benchmarks. Finally, one image optimized jointly for Qwen3.5-4B and Gemma-4-E4B-it improves both models, allowing the image to transfer between the two. In summary, our contributions are:
\begin{itemize}[nosep,leftmargin=*]
    \item The introduction of \textsc{Skill.png}, a gradient-based agent skill insertable via the input space of a frozen VLM, which requires neither weight changes nor white-box access at deployment;
    \item an ensemble objective with per-model gradient normalization that optimizes a transferable image for VLMs of different families;
    \item evaluation of image skills against LLM-optimized text skills on four VLMs and four benchmarks, covering multimodal skill combinations, transfer across models, and an interpretation attempt of image skills.
\end{itemize}
\section{Related Work}
\label{sec:related_work}
Our approach combines image optimization via frozen VLMs with agent skills.

\textbf{Agent skills.} Skills are reusable natural language instructions and resources that help agents perform specific tasks, use tools, and adapt to new domains. Beyond manually writing skills, obtaining skills more systematically is comparable to prompt optimization \cite{Fernando-Promptbreeder,zhou2023large,yang2024large}, where large language models (LLMs) are tasked with providing better instructions. Work in skill optimization has therefore developed increasingly complex pipelines to deploy LLM optimizers for skill writing \cite{agrawal2026gepa,alzubi2026evoskill,ni-etal-2026-trace2skill,yang2024large,zhou2026mementoskillsletagentsdesign}. However, all of these approaches stick to text editing and inefficient sampling from LLMs.

\textbf{Image optimization.} A new line of work has begun to explore multimodal skills. These works typically develop a combination of text and screenshots or visual highlighting that guides agents like an illustrated manual \cite{jiang2026visualskillmultimodalskillscomputeruse,xu2026agentskillstextcase}. However, they rely on manual or only partially automated selections of suitable illustrations. Generating image perturbations through gradient-based updates is therefore much closer to visual prompting \cite{bahng2022exploringvisualpromptsadapting,enomoto2025enhancing}, which has mostly been used in AI safety contexts to jailbreak VLMs \cite{carlini,Bailey-Image-Hijacks,Wang-White-Box-MM-Jailbreak,Zhang-Self-interpr2025,aichberger2026mipagentmaliciousimage,cui2026toward, dobre2026visual}. In contrast to this work, we attempt to elicit a more general behavioral effect through images that goes beyond the narrow target of producing a single adversarial output from VLMs. Moreover, this behavior is intended to improve agents instead of breaking them.
Closest in spirit to our work is ART \cite{chudoba2026finetuningmultimodalllmsart}, which also optimizes an image through a frozen VLM to increase task performance; we instead consider the image as an agent skill, compare it with optimized text skills, combine both, and optimize one image jointly for VLMs of different families.
\section{Method}
\label{sec:method}

Let \(f_\vartheta: \mathcal{X} \rightarrow \mathcal{Y}\) denote a vision-language model (VLM) with trainable parameters \(\vartheta\). To optimize an image skill, \(f_\vartheta\) is provided with the image \(I_{\text{skill}}\), text \(T_{\text{text}}\) (i.e. system message, user request, and previous tool outputs), and target answer \(T_{\text{target}}\) as input:
\begin{equation}
    X_{\text{input}} = \big[\, T_{\text{text}}, I_{\text{skill}}, T_{\text{target}}\big],
\end{equation}
where
\begin{equation}
    I_{\text{skill}} \in [0,1]^{\,3\times H\times W}
\end{equation}
represents an image with three color channels, height \(H\), and width \(W\).

Similar to supervised finetuning of language models, supervision is applied only to the target span \(T_{\text{target}}\), which consists of tokens \(y_1,\dots,y_T\), using causal masking. For each target position \(t \in [T]\), the model produces next-token logits
\begin{equation}
    z_t = f_\vartheta\left(
    T_{\text{text}},\allowbreak
    I_{\text{skill}},\allowbreak
    y_1,\dots,y_{t-1}
    \right) \in \mathbb{R}^{|V|}
\end{equation}
over the vocabulary \(V\). Given these logits, the token-level cross-entropy loss is defined as
\begin{equation}
      \ell_t = -\log \frac{\exp\!\big(z_{t, y_t}\big)}
                   {\sum_{v \in V} \exp\!\big(z_{t, v}\big)} ,
\end{equation}
which forms the basis of our training objective.
    
With all target tokens weighted equally, the objective becomes the mean cross-entropy over the target tokens and can be rewritten as the loss,
\begin{equation}
    \mathcal{L}_{\text{CE}}
    = \frac{1}{T} \sum_{t=1}^{T} \ell_t .
\end{equation}
During optimization, the model's parameters $\vartheta$ are frozen; gradients are taken only with respect to $I_{\text{skill}}$ from each training target and propagate through language and vision modules of $f_\vartheta$ to the input. After each gradient step, $I_{\text{skill}}$ is projected back onto the valid image domain $[0,1]^{\,3\times H\times W}$ by element-wise clamping.

\subsection{Ensemble Optimization}
\label{sec:ensemble_objective}

To obtain a transferable skill in pixel space, we optimize one shared $I_{\text{skill}}$ jointly against an ensemble of $K$ frozen VLMs $f_{\vartheta_1},\dots,f_{\vartheta_K}$. Each model receives the same text, image and target, processes $I_{\text{skill}}$ with its own vision module, and yields its own cross-entropy loss $\mathcal{L}_{\text{CE}}^{(k)}$ as defined above.

The ensemble objective is the weighted mean of the per-model losses,
\begin{equation}
    \mathcal{L}_{\text{ens}}
    = \sum_{k=1}^{K} w_k \, \mathcal{L}_{\text{CE}}^{(k)},
    \qquad w_k > 0, \quad \sum_{k=1}^{K} w_k = 1 .
\end{equation}
Normalizing the weights keeps $\mathcal{L}_{\text{ens}}$ on the same scale as a single-model loss. We use uniform weights, $w_k = 1/K$.

The gradient of $\mathcal{L}_{\text{ens}}$ is the weighted sum of the per-model gradients
$g_k = \nabla_{I_{\text{skill}}} \mathcal{L}_{\text{CE}}^{(k)}$. Their magnitudes can differ considerably between models, for example, due to different loss levels or vision modules, so a single VLM can dominate the update. We therefore rescale each gradient to the unit $L_2$ norm before combining them:
\begin{equation}
    \tilde{g} = \sum_{k=1}^{K} w_k \, \frac{g_k}{\lVert g_k \rVert_2} .
\end{equation}
The optimizer step uses $\tilde{g}$ in place of $\nabla_{I_{\text{skill}}} \mathcal{L}_{\text{ens}}$. Each VLM therefore pulls the image in its own direction with strength $w_k$, whatever its gradient scale. As in the single-model case, $I_{\text{skill}}$ is clamped to $[0,1]^{\,3\times H\times W}$ after each step.

\section{Experimental Setup}
\label{sec:experiments}

% Non-reasoning: main results incl. Random Image, LLM-Skill + Skill.png, LoRA reference rows, plus SkillOpt (Sonnet 5.5) and Synergistic on Nemotron-3
\begin{table*}[t]
    \centering
    {\footnotesize
    \begin{tabular}{llllll}
    \toprule
    Model & Skill source & SearchQA & TRAJECT & LiveMath & T1 \\
    \midrule
    Qwen3.5-4B & No Skill & 68.79 & 25.37 & 48.54 & 24.71 \\
     & Random Image & 71.00\gain{2.21} & 25.20\loss{0.16} & 42.68\loss{5.86} & 24.79\gain{0.07} \\
     & LLM-Skill & 71.00\gain{2.21} & \underline{27.66}\gain{2.29} & 49.37\gain{0.84} & 25.64\gain{0.93} \\
     & SkillOpt & 71.07\gain{2.29} & \textbf{32.08\gain{\textbf{6.71}}} & 46.03\loss{2.51} & \underline{27.86}\gain{3.14} \\
     & \cellcolor{skillbg}\textbf{Skill.png} & \cellcolor{skillbg}\textbf{75.36\gain{\textbf{6.57}}} & \cellcolor{skillbg}25.20\loss{0.16} & \cellcolor{skillbg}\textbf{62.34\gain{\textbf{13.81}}} & \cellcolor{skillbg}\textbf{50.50\gain{\textbf{25.79}}} \\
     & \cellcolor{skillbg}LLM-Skill + Skill.png & \cellcolor{skillbg}\underline{73.50}\gain{4.71} & \cellcolor{skillbg}27.33\gain{1.96} & \cellcolor{skillbg}\underline{50.21}\gain{1.67} & \cellcolor{skillbg}25.21\gain{0.50} \\
    \cmidrule(l){2-6}
     & \cellcolor{gray!12}LoRA \tiny latent space & \cellcolor{gray!12}80.71\gain{11.93} & \cellcolor{gray!12}56.30\gain{30.93} & \cellcolor{gray!12}80.33\gain{31.80} & \cellcolor{gray!12}80.93\gain{56.21} \\
    \midrule
    Gemma-4-E4B-it & No Skill & 61.64 & 40.26 & 49.79 & 24.71 \\
     & Random Image & 61.71\gain{0.07} & 35.84\loss{4.42} & \underline{51.05}\gain{1.26} & 24.71\textsubscript{$\pm$0.00} \\
     & LLM-Skill & 68.07\gain{6.43} & \textbf{41.73\gain{\textbf{1.47}}} & 41.84\loss{7.95} & 25.07\gain{0.36} \\
     & SkillOpt & 69.36\gain{7.71} & \underline{41.57}\gain{1.31} & 49.79$^*$\textsubscript{$\pm$0.00} & \underline{41.21}\gain{16.50} \\
     & \cellcolor{skillbg}\textbf{Skill.png} & \cellcolor{skillbg}\underline{71.93}\gain{10.29} & \cellcolor{skillbg}40.26$^*$\textsubscript{$\pm$0.00} & \cellcolor{skillbg}\textbf{55.23\gain{\textbf{5.44}}} & \cellcolor{skillbg}\textbf{62.50\gain{\textbf{37.79}}} \\
     & \cellcolor{skillbg}LLM-Skill + Skill.png & \cellcolor{skillbg}\textbf{72.79\gain{\textbf{11.14}}} & \cellcolor{skillbg}\textbf{41.73$^*$\gain{\textbf{1.47}}} & \cellcolor{skillbg}44.77\loss{5.02} & \cellcolor{skillbg}36.36\gain{11.64} \\
    \cmidrule(l){2-6}
     & \cellcolor{gray!12}LoRA \tiny latent space & \cellcolor{gray!12}77.57\gain{15.93} & \cellcolor{gray!12}53.52\gain{13.26} & \cellcolor{gray!12}75.73\gain{25.94} & \cellcolor{gray!12}83.21\gain{58.50} \\
    \midrule
    Qwen3.5-9B & No Skill & 71.07 & 47.79 & 46.03 & 27.21 \\
     & Random Image & 71.07\textsubscript{$\pm$0.00} & 49.26\gain{1.47} & 39.33\loss{6.69} & 27.14\loss{0.07} \\
     & LLM-Skill & \textbf{74.71\gain{\textbf{3.64}}} & \underline{50.25}\gain{2.45} & \underline{49.79}\gain{3.77} & 28.00\gain{0.79} \\
     & SkillOpt & \underline{73.79}\gain{2.71} & \textbf{51.39\gain{\textbf{3.60}}} & 46.03$^*$\textsubscript{$\pm$0.00} & \underline{44.29}\gain{17.07} \\
     & \cellcolor{skillbg}\textbf{Skill.png} & \cellcolor{skillbg}72.14\gain{1.07} & \cellcolor{skillbg}48.12\gain{0.33} & \cellcolor{skillbg}\textbf{58.16\gain{\textbf{12.13}}} & \cellcolor{skillbg}\textbf{44.57\gain{\textbf{17.36}}} \\
     & \cellcolor{skillbg}LLM-Skill + Skill.png & \cellcolor{skillbg}73.29\gain{2.21} & \cellcolor{skillbg}49.75\gain{1.96} & \cellcolor{skillbg}48.12\gain{2.09} & \cellcolor{skillbg}28.36\gain{1.14} \\
    \cmidrule(l){2-6}
     & \cellcolor{gray!12}LoRA \tiny latent space & \cellcolor{gray!12}82.86\gain{11.79} & \cellcolor{gray!12}51.55\gain{3.76} & \cellcolor{gray!12}80.33\gain{34.31} & \cellcolor{gray!12}83.79\gain{56.57} \\
    \midrule
    Nemotron-3-Nano- & No Skill & 65.29 & 30.93 & 54.39 & 24.79 \\
    Omni-30B-A3B & Random Image & 66.93\gain{1.64} & \textbf{33.88\gain{\textbf{2.95}}} & 52.30\loss{2.09} & 24.79\textsubscript{$\pm$0.00} \\
     & LLM-Skill & 68.86\gain{3.57} & 28.97\loss{1.96} & 55.65\gain{1.26} & 25.07\gain{0.29} \\
     & SkillOpt & 68.00\gain{2.71} & \underline{31.26}\gain{0.33} & 57.32\gain{2.93} & 26.86\gain{2.07} \\
     & \cellcolor{skillbg}\textbf{Skill.png} & \cellcolor{skillbg}\textbf{75.79\gain{\textbf{10.50}}} & \cellcolor{skillbg}21.11\loss{9.82} & \cellcolor{skillbg}\underline{70.71}\gain{16.32} & \cellcolor{skillbg}\textbf{55.36\gain{\textbf{30.57}}} \\
     & \cellcolor{skillbg}LLM-Skill + Skill.png & \cellcolor{skillbg}\underline{74.21}\gain{8.93} & \cellcolor{skillbg}21.44\loss{9.49} & \cellcolor{skillbg}\textbf{72.38\gain{\textbf{17.99}}} & \cellcolor{skillbg}\underline{35.29}\gain{10.50} \\
    \cmidrule(l){2-6}
     & \cellcolor{gray!12}LoRA \tiny latent space & \cellcolor{gray!12}81.57\gain{16.29} & \cellcolor{gray!12}39.44\gain{8.51} & \cellcolor{gray!12}82.43\gain{28.03} & \cellcolor{gray!12}77.43\gain{52.64} \\
    \cmidrule(l){2-6}
     & SkillOpt (Sonnet 5.5) & 69.07\gain{3.79} & 40.59\gain{9.66} & 60.25\gain{5.86} & 43.43\gain{18.64} \\
     & \cellcolor{skillbg}\textbf{Joint Optim.}& \cellcolor{skillbg}75.21\gain{9.93} & \cellcolor{skillbg}40.59$^\dagger$\gain{9.66} & \cellcolor{skillbg}75.73\gain{21.34} & \cellcolor{skillbg}58.36\gain{33.57} \\

    \bottomrule
    \end{tabular}}
\caption{\textbf{Test scores of text and image skills.} Random Image is an unoptimized uniform-noise image of each model's \textsc{Skill.png} size, LLM-Skill is written by GPT-5.5 in a single pass, SkillOpt is refined by GPT-6 Luna from the target model's rollouts, and \textsc{Skill.png} is an image optimized by gradient descent through the frozen target model. LLM-Skill + \textsc{Skill.png} adds the LLM-Skill to the system prompt and the \textsc{Skill.png} image, which was optimized without the skill, to the user turn. LoRA {\tiny latent space}, shown for reference. Scores in \%, with the change from No Skill in percentage points as subscript. Best score per model and benchmark among the first six rows of each model in bold, second best underlined; ties share a mark. $^*$Optimization found nothing better than No Skill on validation, so the score without the optimized skill or image is shown (No Skill, or LLM-Skill for the combination). \textbf{Jointly optimized skills on Nemotron-3}: SkillOpt (Sonnet 5.5) swaps the optimizer LLM, and Joint Optim. adds an image that was trained and tested with the Sonnet 5.5 skill in the system prompt. This combination gives the highest LiveMath and T1 scores, above either skill alone. $^\dagger$No image beat the Sonnet 5.5 skill alone on validation, so the skill's score is shown.}
\label{tab:main-non-reasoning}
\end{table*}

\subsection{Benchmarks}
\label{sec:benchmarks}
We evaluate on four benchmarks. SearchQA \cite{dunn2017searchqanewqadataset} poses \textit{Jeopardy!} questions together with web snippets retrieved for clues. The agent's answers are scored by exact match. TRAJECT-Bench \cite{he2026trajectbencha} is a tool use benchmark for everyday tasks. The agent must emit every required tool call from a set of tools that also contains 10 distractors. An answer counts as correct only if both the selected tools and their required parameters match exactly. LiveMathematicianBench (LiveMath) \cite{he2026livemathematicianbench} asks five-way multiple-choice questions on theorems from recent arXiv papers, scored by accuracy. T1 \cite{Chakraborty_T1_benchmark} covers multi-turn travel planning: at a user turn, the agent writes Python code against flight, hotel, restaurant and attraction tools. The code is executed and scored by exact match of the resulting benchmark cache states. All benchmarks use fixed train/validation/test splits, and all reported scores are on our test split. We share further benchmark details in the Supplementary Material. 

\subsection{Vision-Language Models}
\label{sec:VLMs}
We optimize skill images for four VLMs: Qwen3.5-4B \cite{qwen35blog}, Qwen3.5-9B \cite{qwen35blog}, Gemma-4-E4B-it \cite{gemmateam2026gemma4} and Nemotron-3-Nano-Omni-30B-A3B (Nemotron-3) \cite{nvidia2026nemotron3nanoomni}, all kept frozen in bfloat16. Images are optimized with reasoning disabled, and unless stated otherwise, models are also evaluated in non-reasoning mode.

A VLM's preprocessor resizes any image that does not match its patch grid or token budget, which would place a resampling step between $I_{\text{skill}}$ and the vision encoder. We therefore set $H \times W$ per model to the smallest square size that its preprocessor passes through unchanged, so that every optimized pixel reaches the encoder as is: $256 \times 256$ for Qwen3.5, the minimum pixel count of its dynamic resizing; $512 \times 512$ for Nemotron-3, below which its preprocessor upscales to a minimum of 1024 patches; and $768 \times 768$ for Gemma-4-E4B-it, to which its preprocessor maps every square image to fill a fixed soft-token budget. This gives 64, 256 and 256 image tokens, for each VLM respectively.

$I_{\text{skill}}$ is initialized to uniform mid-gray and then optimized with Adam for 80 epochs over the training split, with a batch size of 12 and a learning rate of 0.03 that decays to zero on a cosine schedule. We lower the momentum term to $\beta_1 = 0.5$ ($\beta_2 = 0.999$) so that each update follows the current gradient more closely (consider our ablation on $\beta_1$ in Sec.~\ref{sec:ablations}). Every 100 steps, the image is scored on the validation split with the benchmark metric, and the best-scoring image is kept for testing. If no image beats the model without a skill on validation, we report the No Skill score, marked with $^*$ in the tables.

\subsection{Baselines}
\label{sec:baselines}
\looseness=-1
LLM-Skill is a text skill that GPT-5.5 writes in a single pass from a description of the benchmark, its system prompt and five training examples, without feedback from any target model. One such skill per benchmark is therefore shared by all models. SkillOpt \cite{yang2026skilloptexecutivestrategyselfevolving} creates a text skill from scratch, with GPT-6 Luna -- as a cheaper alternative to GPT-5.5 -- reflecting on each target model's rollouts on the training split and keeping only edits that raise the validation score. Both skills are appended to the system prompt. Because none of the VLMs we tested allows images in the system prompt, all images must be injected with the first user request. The Supplementary Material lists all hyperparameters of our text-based optimizations as well as details of supervised finetuning via LoRA \cite{hu2022lora} as another reference.

\subsection{Hybrid Skill Experiments}
To test whether skills from different modalities are synergistic, we inject skills from LLM-Skill together with \textsc{Skill.png} images at test time. In addition to this experiment, we run SkillOpt on Nemotron-3 with Claude Sonnet 5.5 as the optimizer and then optimize $I_{\text{skill}}$ with the text skill appended to the system prompt. The SkillOpt skill is thus part of $T_{\text{text}}$ during the training, validation and test query.
\section{Main Results}
\label{sec:main_results}

\noindent\textbf{Test scores.} \textsc{Skill.png} significantly outperforms \mbox{SkillOpt} in 9 of 16 model--benchmark pairs (paired exact McNemar test, $p<0.05$; Tab.~\ref{tab:main-non-reasoning}), with the largest gains on T1\footnote{In 345 of the 1400 T1 test turns (24.64\%), the reference solution leaves the benchmark cache unchanged, such that any response is scored as correct, explaining the similar baseline scores of all tested VLMs.} ($+17.36$ to $+37.79$ points over No Skill) and LiveMath ($+5.44$ to $+16.32$). In two further pairs it
scores higher, but not significantly so: Qwen3.5-9B on T1 ($+0.28$) and
Gemma-4-E4B-it on LiveMath ($+5.44$). It does not beat SkillOpt on any
TRAJECT pair or for Qwen3.5-9B on SearchQA. On TRAJECT, the image does not even improve significantly over No Skill and drops by 9.82 points on Nemotron-3. LiveMath mirrors this for SkillOpt: it gains only on Nemotron-3 ($+2.93$), finds no skill for Gemma-4-E4B-it and Qwen3.5-9B, and loses 2.51 points on Qwen3.5-4B. On Nemotron-3, replacing GPT-6 Luna with Claude Sonnet 5.5 as the \mbox{SkillOpt} optimizer raises the text skill on all four benchmarks, most on T1 ($+16.57$) and TRAJECT ($+9.33$). \textsc{Skill.png} still beats this stronger text skill on SearchQA, LiveMath and T1, by 6.71 to 11.93 points. LoRA, which is confined to the latent space and does not produce any skill artifacts in the input space, would raise the models the most on all benchmarks.

\noindent\textbf{Skill synergy.} Adding the LLM-Skill to the system prompt of an image optimized without it does not combine their score gains: the combination beats both of its parts in only 2 of 16 model--benchmark cells, by at most 1.67 points, and falls 16.21 to 26.14 points below the image alone on T1. In contrast, an image optimized with the Claude Sonnet 5.5 skill already in the system prompt adds 6.14 to 15.48 points to that text skill on SearchQA, LiveMath and T1. These jointly optimized skills set the best Nemotron-3 scores on LiveMath (75.73) and T1 (58.36), 5.02 and 3.00 points above the image alone and match \textsc{Skill.png} on SearchQA.

\begin{figure}[t]
  \centering
  \includegraphics[width=\linewidth]{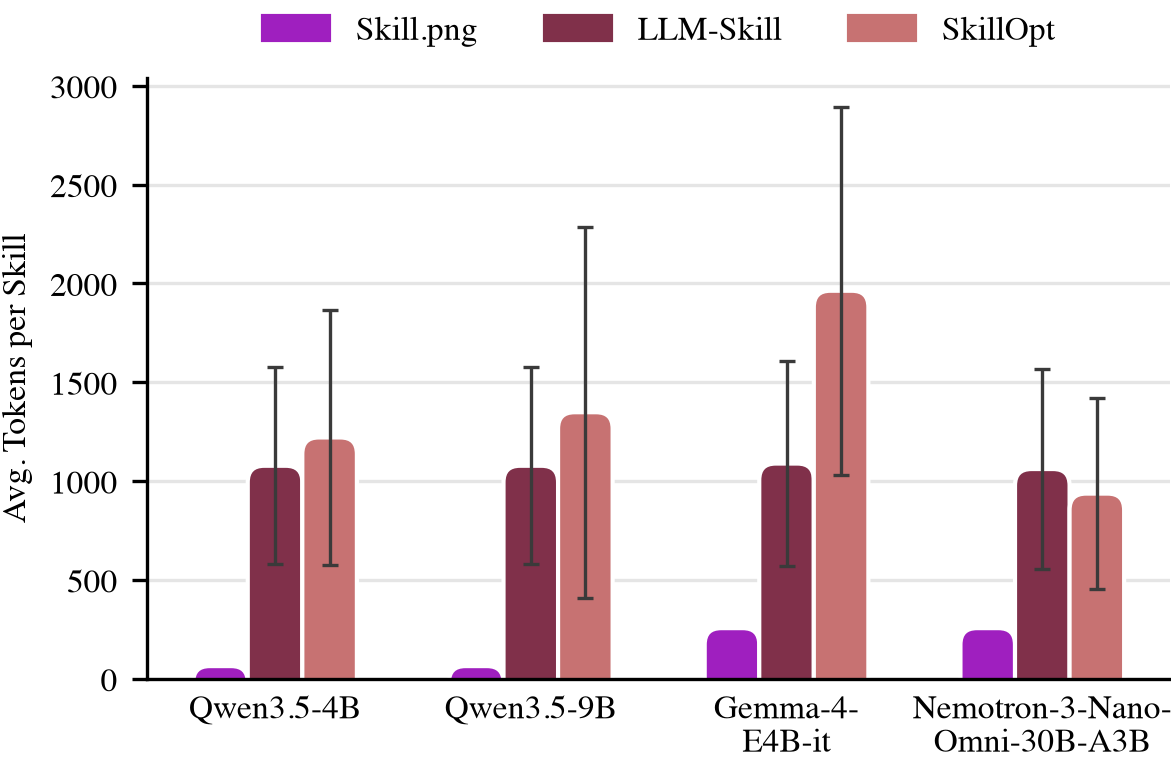}
  \caption{\textbf{Length of text and image skills.} Tokens per skill and model, averaged over the four benchmarks, with error bars showing one standard deviation across them. For \textsc{Skill.png}, these are image tokens, which are fixed per model and so have no error bar.}
  \label{fig:token_count}
\end{figure}

\begin{table}[t]
  \centering
  {\footnotesize
  \begin{tabular}{lrr}
    \toprule
    Model & SkillOpt & Skill.png \\
    \midrule
    Qwen3.5-4B & 8.70M in / 0.75M out & 12.20 GPU h \\
    Qwen3.5-9B & 8.84M in / 0.64M out & 19.86 GPU h \\
    Gemma4-E4B & 8.80M in / 0.70M out & 16.89 GPU h \\
    Nemotron3-A3B & 8.17M in / 0.72M out & 19.46 GPU h \\
    % Qwen3.5-4B & 8.70M in / 0.75M out & 12.20 GPU h \\
    % Qwen3.5-9B & 8.84M in / 0.64M out & 19.86 GPU h \\
    % Gemma-4-E4B-it & 8.80M in / 0.70M out & 16.89 GPU h \\
    %\begin{tabular}[t]{@{}l@{}}Nemotron-3-Nano-\\Omni-30B-A3B\end{tabular} & 8.17M in / 0.72M out & 19.46 GPU h \\
    \bottomrule
  \end{tabular}}
  \caption{\textbf{Optimization cost of SkillOpt and \textsc{Skill.png}.} Averages over SearchQA, TRAJECT, LiveMath and T1. SkillOpt: input and output tokens of the GPT-6 Luna optimizer. \textsc{Skill.png}: GPU hours of image training on NVIDIA H100 and H200 GPUs, summed over all GPUs a model was split across.}
  \label{tab:cost}
\end{table}

\noindent\textbf{Efficiency.} \textsc{Skill.png} adds 64 (Qwen3.5) or 256 (Gemma-4-E4B-it, Nemotron-3) image tokens per query, whereas the text skills average 937 to 1963 tokens with a high benchmark-dependent deviation (Fig.~\ref{fig:token_count}). Optimizing a skill image takes 12.2 to 19.9 GPU hours per benchmark. SkillOpt needs only model outputs but, on average in our experiments with GPT-6 Luna, spends over 8.8M input and output tokens (Tab.~\ref{tab:cost}). We provide these rough approximations with practitioners in mind; a more fine-grained comparison is not possible due to a lack of information about GPT-6 Luna's properties.

\begin{figure*}[t]
  \centering
  \includegraphics[width=\textwidth]{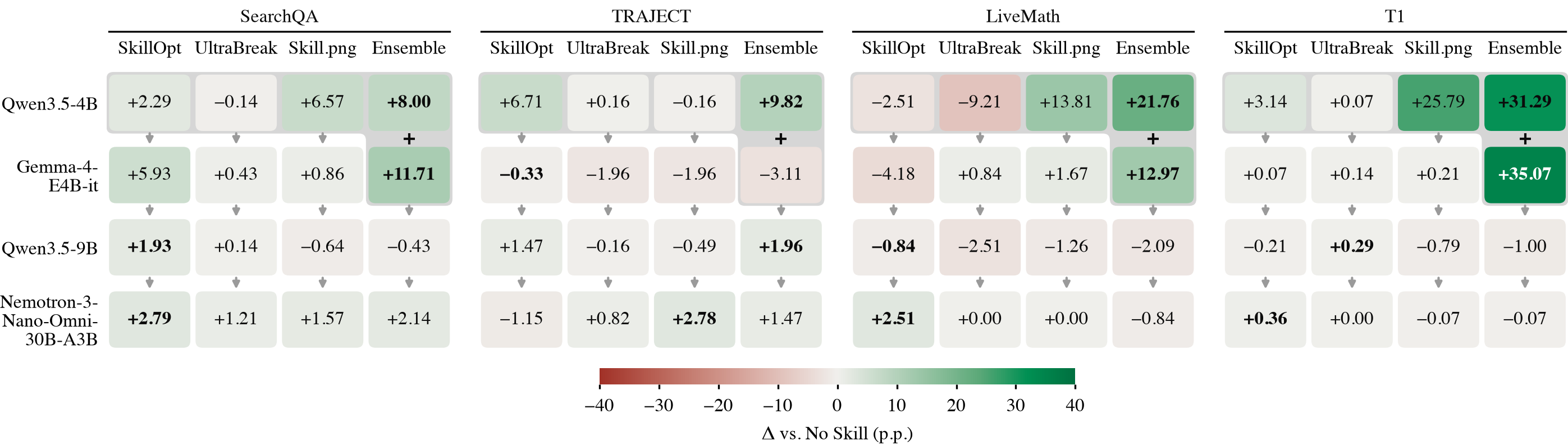}
    \caption{\textbf{Transfer of skills across models.} Change in test score over No Skill, in percentage points, when skills optimized on Qwen3.5-4B are injected unchanged into each model. SkillOpt: text skill optimized by GPT-6 Luna on Qwen3.5-4B. UltraBreak: image optimized on Qwen3.5-4B. \textsc{Skill.png}: image optimized on Qwen3.5-4B alone. Ensemble: image optimized jointly on Qwen3.5-4B and Gemma-4-E4B-it. Gray bands mark the models each skill was optimized on, and arrows point to the models it is injected into. Bold: best skill per model and benchmark.}
  \label{fig:transfer}
\end{figure*}

\subsection{Transferability of Skills}
\label{sec:transferability}

We test transfer with Qwen3.5-4B as the source model and inject each of its skills unchanged into Gemma-4-E4B-it, Qwen3.5-9B and Nemotron-3. We compare four skills: the SkillOpt text skill that GPT-6 Luna optimized on Qwen3.5-4B; an UltraBreak image \citep{cui2026toward} optimized on Qwen3.5-4B; the \textsc{Skill.png} image optimized on Qwen3.5-4B alone; and an ensemble image optimized jointly on Qwen3.5-4B and Gemma-4-E4B-it with the objective of Section~\ref{sec:ensemble_objective}. UltraBreak was built for jailbreak images that transfer across models. We use it as another image optimization as well as a transfer baseline. Its loss is not cross-entropy on the target tokens. Instead, it pulls a token's embedding expected under the model's next-token distribution toward the embeddings of the target tokens. During training, the image is randomly rotated, scaled and shifted, with an additional total-variation penalty. We set its targets to the gold answers, give it the same step budget as \textsc{Skill.png}, and keep its final image, as UltraBreak does. We share all hyperparameters used for UltraBreak in the Supplementary Material. At test time, each transferred image is passed at its optimized size and the target model's own preprocessor resizes it where the target model expects a larger input.

The image optimized on Qwen3.5-4B alone changes performance of Gemma-4-E4B-it by only $-1.96$ to $+1.67$ points, whereas the ensemble image raises it by 11.71 to 35.07 points on SearchQA, LiveMath and T1 ($p < 0.01$). On Qwen3.5-4B, the ensemble image even surpasses the model's own \textsc{Skill.png} on all four benchmarks, significantly on TRAJECT ($+9.98$), LiveMath ($+7.95$) and T1 ($+5.50$) (Fig.~\ref{fig:transfer}). On TRAJECT, the ensemble image also lifts Qwen3.5-4B by 9.82 points, while no model's own image improves over No Skill (Tab.~\ref{tab:main-non-reasoning}).

However, transfer beyond the ensemble remains weak for all kinds of skills. The SkillOpt skill from GPT-6 Luna of Qwen3.5-4B raises all three target models on SearchQA by 1.93 to 5.93 points ($p < 0.05$), but no other benchmark significantly. The \textsc{Skill.png} from Qwen3.5-4B changes the held-out Qwen3.5-9B and Nemotron-3 by $-2.09$ to $+2.78$ points, and none of the ensemble images improve transfer to unseen models. UltraBreak does not work off the shelf for skills: its image does not even improve Qwen3.5-4B, on which it is optimized, and lowers LiveMath by 9.21 points ($p < 0.05$). Consequently, it does not transfer either, changing the other three models by $-2.51$ to $+1.21$ points. The few significant gains, all on Nemotron-3, are on par with those of a random image (compare with Tab.~\ref{tab:main-non-reasoning}). We, therefore, report that joint optimization extends an image to the models it is optimized on, but not to unseen ones.

Beyond other models, we also test transfer to reasoning mode: each skill, optimized with reasoning disabled, is injected unchanged into the same model with reasoning enabled. We test SearchQA, TRAJECT and T1. \textsc{Skill.png} still improves on No Skill in reasoning mode in 7 of these 12 settings, but beats the SkillOpt skill less consistently (see the Supplementary Material for detailed results). 

\subsection{Ablations}
\label{sec:ablations}

We vary the size of the skill image for Qwen3.5-4B on SearchQA and T1, from $64\times64$ to $768\times768$, and keep the rest of the recipe fixed. Sizes below $256\times256$ are bicubically upscaled to it by Qwen3.5-4B's preprocessor before they reach the encoder, so they keep the token count and only reduce the number of free pixels. Up to $320\times320$, we further optimize each image with three training seeds and report the mean and sample standard deviation across seeds. Because they are the most expensive to optimize, the $512\times512$ and $768\times768$ images use a single seed. To test whether the gains come from the optimized content rather than from the presence of an image, we also replace \textsc{Skill.png} with a uniform-noise image of the same size for each model.

\begin{figure}[t]
  \centering
    \includegraphics[width=\columnwidth]
            {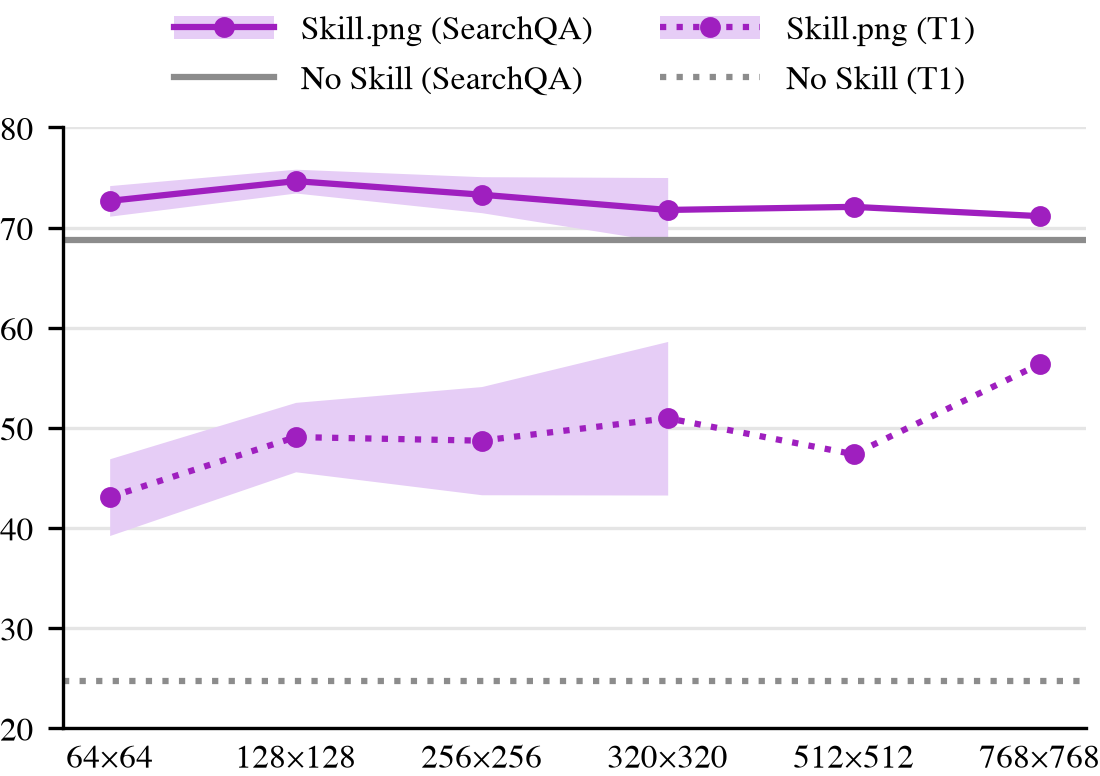}
    \captionof{figure}{\textbf{Effect of different image resolutions.} Test score (\%) of Qwen3.5-4B with \textsc{Skill.png} images optimized at six resolution sizes. Up to $320\times320$, points show the mean over three training seeds and shaded bands one sample standard deviation; $512\times512$ and $768\times768$ use a single seed.}
    \label{fig:resolution}
\end{figure}

\looseness=-1
Every image size improves over No Skill on both benchmarks (Fig.~\ref{fig:resolution}), but larger images do not help consistently. For the sizes with three training seeds, the standard deviation is 1.19 to 3.24 points on SearchQA and 3.47 to 7.67 points on T1. On SearchQA, scores peak at $128\times128$ ($74.64\pm1.19$); at $320\times320$ ($71.76\pm3.24$), one standard deviation reaches down to No Skill (68.79). On T1, scores rise from $43.07\pm3.83$ at $64\times64$ to $50.95\pm7.67$ at $320\times320$; an image of $768\times768$ lifts scores to 56.36 while using $5.76\times$ the image tokens of $320\times320$ (576 vs. 100), a gain that stays within one standard deviation of $320\times320$. This being said, the seeds of a single size differ by up to 14.00 points, so the reported results at $512\times512$ and $768\times768$ should be read with caution. The $64\times64$ canvas, with $1/16$ of the free pixels at the same 64 tokens, keeps 86.24\% of the $256\times256$ gain on SearchQA and 76.49\% on T1. Therefore, it appears that most of the skill information fits into a few pixels.

The noise control (consider Random Image in Tab.~\ref{tab:main-non-reasoning}) shows that the gains come from the optimized content, not from the presence of an image. Over all models and benchmarks, the noise image changes the score by between $-6.69$ and $+2.95$ points, and \textsc{Skill.png} beats it on all twelve model-benchmark pairs of SearchQA, LiveMath and T1. On T1, noise leaves every model within $0.07$ points of No Skill, while \textsc{Skill.png} adds $17.36$ to $37.79$ points. On LiveMath, noise lowers the score of three of the four models, by up to $6.69$ points. TRAJECT is the exception: neither image helps there, and noise even scores above \textsc{Skill.png} on Qwen3.5-9B and Nemotron-3, in line with the overall missing \textsc{Skill.png} gains on TRAJECT.

\begin{figure}[t]
  \centering
    \includegraphics[width=\columnwidth]
            {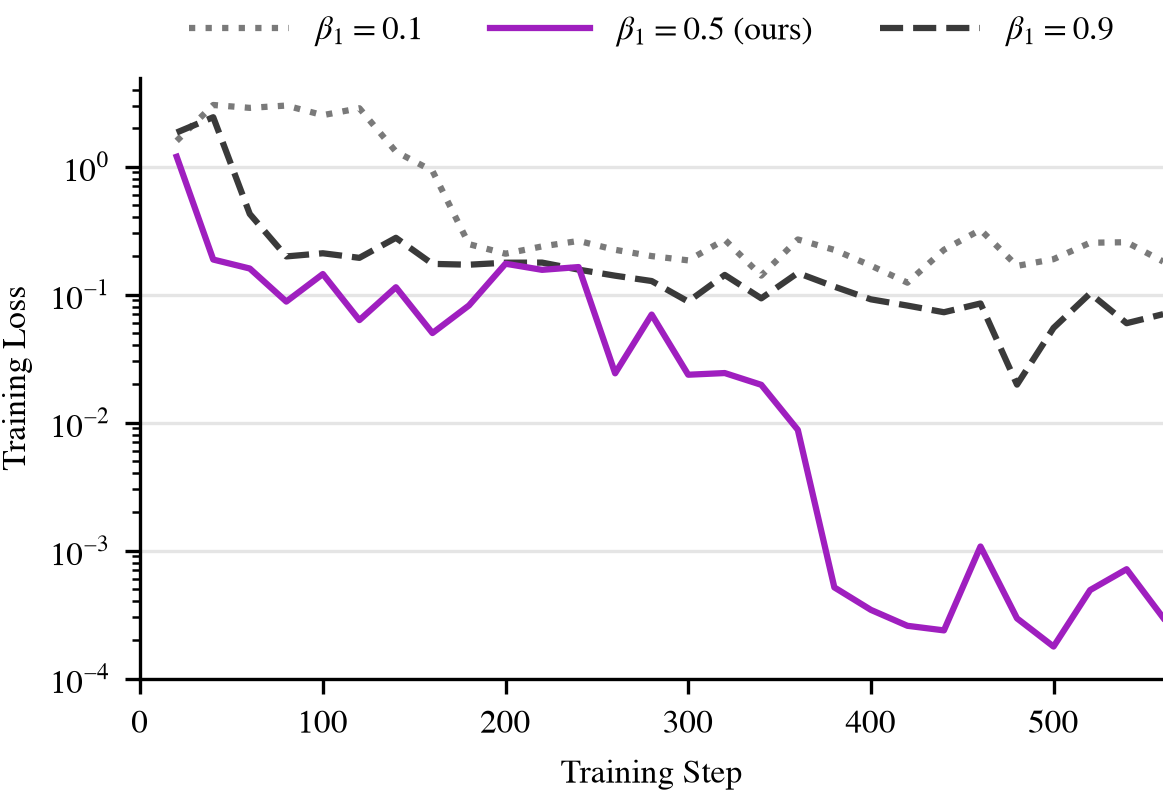}
    \captionof{figure}{\textbf{Effect of Adam momentum.} Training loss (log scale) of \textsc{Skill.png} for Qwen3.5-9B on LiveMath with $\beta_1 \in \{0.1, 0.5, 0.9\}$. Each point is the loss on one 12-item batch, logged every 20 steps.}
    \label{fig:momentum}
\end{figure}

We also ablate Adam's momentum coefficient $\beta_1$, which sets how much of the past gradients each update carries. We train \textsc{Skill.png} for Qwen3.5-9B on LiveMath with $\beta_1 \in \{0.1, 0.5, 0.9\}$, keeping $\beta_2 = 0.999$ and all other hyperparameters fixed (Fig.~\ref{fig:momentum}). With $\beta_1 = 0.5$ the training loss falls fastest and ends more than two orders of magnitude below the other values ($3\times10^{-4}$ vs.\ $7\times10^{-2}$ for PyTorch's default $0.9$). $\beta_1 = 0.5$ also scores best on the test set (58.16 vs.\ 48.95 for $\beta_1 = 0.9$ and 42.68 for $\beta_1 = 0.1$; 46.03 without a skill). We therefore use $\beta_1 = 0.5$ in all other experiments.

\subsection{Interpreting Skill.png}
\label{sec:interpretability}

\begin{figure}[t]
  \centering
    \begin{minipage}[t]{0.48\linewidth}\centering
      \includegraphics[width=\linewidth]{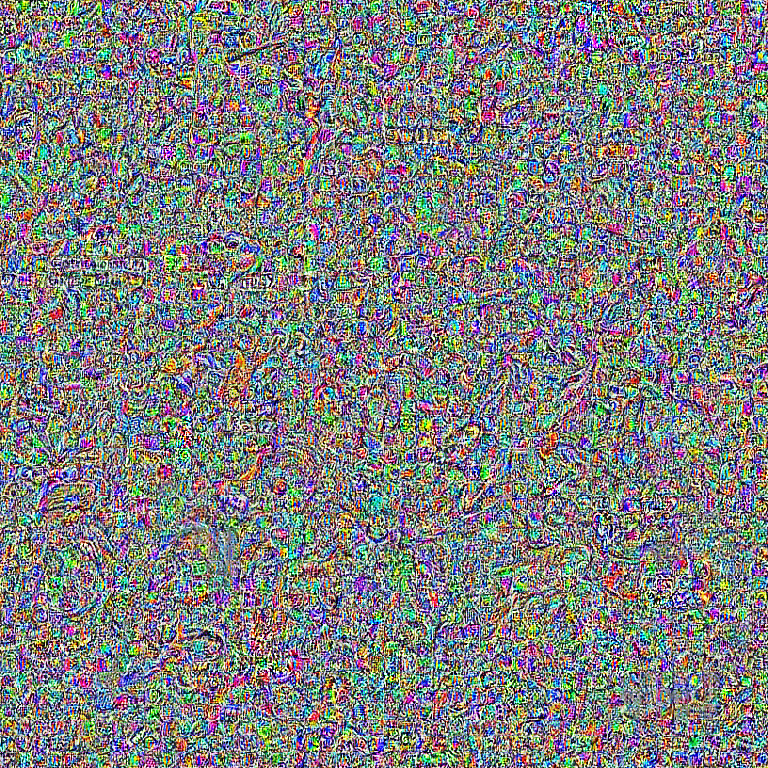}\\[-1pt]
      {\footnotesize SearchQA, $+10.3$}
    \end{minipage}\hfill
    \begin{minipage}[t]{0.48\linewidth}\centering
      \includegraphics[width=\linewidth]{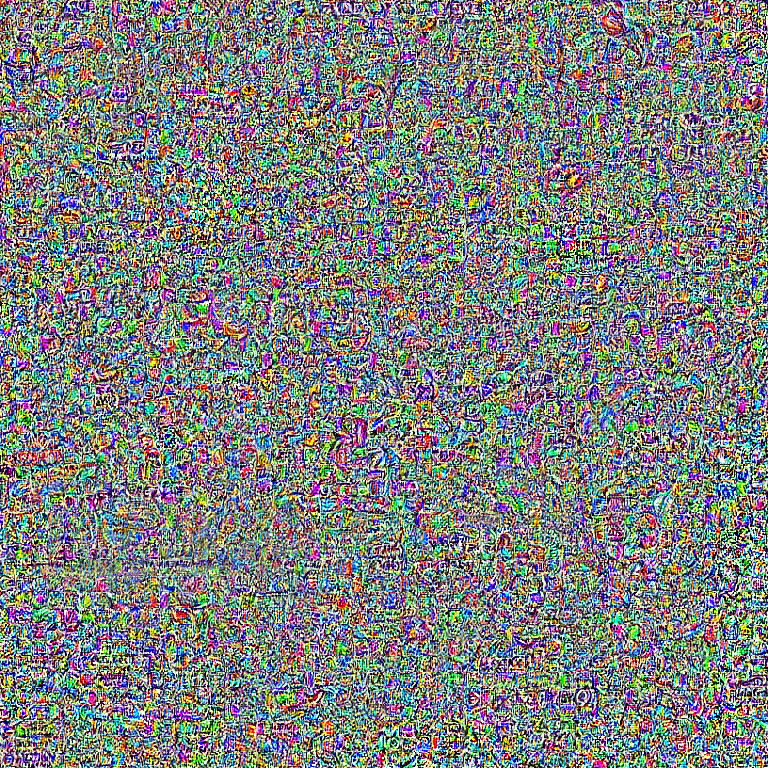}\\[-1pt]
      {\footnotesize T1, $+37.8$}
    \end{minipage}
    \begin{tcolorbox}[probebox]
    \textbf{SearchQA}, \textit{[\ldots] What is it telling you to do?}\\[1pt]
    \textbf{Gemma-4-E4B-it:} ``The image is a fill-in-the-blank quiz or puzzle. It is telling you to fill in the missing words [\ldots]''\\[1pt]
    \textbf{Qwen3.5-9B:} ``[\ldots] It's essentially visual static or `noise' [\ldots]''
    \par\vspace{3pt}{\color{skillbg!80!black}\hrule}\vspace{3pt}
    \textbf{T1}, \textit{Transcribe any text, letters, numbers or symbols you can read in this image [\ldots]}\\[1pt]
    \textbf{Gemma-4-E4B-it:} ``FIND A SWEET TRAVEL [\ldots] in the code snippet: \texttt{explore\_all} \texttt{next\_location} \texttt{start\_location} [\ldots]''\\[1pt]
    \textbf{Qwen3.5-9B:} ``No text is visible in the provided image. The image appears to be a highly distorted, colorful static pattern [\ldots]''
    \end{tcolorbox}
    \captionof{figure}{\textbf{Answers on two \textsc{Skill.png} images of Gemma-4-E4B-it.} Images optimized on SearchQA and T1 ($768\times768$ pixels; gain over No Skill in points), shown without system prompt. One shortened answer per model.}
    \label{fig:gemma_probe}
\end{figure}

A human can read a text skill and check what it tells the agent. A \textsc{Skill.png} image offers no such access: it resembles colorful noise (see examples in Fig.~\ref{fig:gemma_probe}). Its performance gains alone do not reveal whether the VLM extracts content from the image, as it would from a text skill, or whether the pixels merely steer the model without any meaning attached. We therefore test whether \textsc{Skill.png} images carry meaning that the VLMs themselves can put into words. We show every \textsc{Skill.png} image to all four models, together with the random image of our noise control. The image is shared in a user turn with no system prompt and no task, together with one of five open questions: (1) describe the image, (2) transcribe its text, (3) say what instruction it gives, (4) rewrite it as a system prompt, and (5) list ten words it brings to mind. We sample five answers per question \cite{scholten2025probabilistic}. An LLM judge, GLM-5.3-Flash, which sees only the question and the answer, labels each answer, for example whether it calls the image noise or corrupted and which task it connects the image with. We validate its labels with Claude Opus 5.5 as a second judge \cite{schwinn2026coin}; the two judges agree with Cohen's $\kappa = 0.91$ (details are shared in the Supplementary Material).

% Probe: share of answers calling an image noise or corrupted
\begin{table}[t]
    \centering
    {\footnotesize
    \setlength{\tabcolsep}{4pt}
    \begin{tabular}{lrrrr}
    \toprule
     & \multicolumn{4}{c}{Shown to} \\
    \cmidrule(lr){2-5}
    Optimized for & \begin{tabular}[b]{@{}r@{}}Qwen3.5-\\4B\end{tabular} & \begin{tabular}[b]{@{}r@{}}Qwen3.5-\\9B\end{tabular} & \begin{tabular}[b]{@{}r@{}}Gemma-4-\\E4B-it\end{tabular} & \begin{tabular}[b]{@{}r@{}}Nemotron-\\3\end{tabular} \\
    \midrule
    Qwen3.5-4B & \cellcolor{skillbg}\textbf{8\%} & 54\% & 80\% & 47\% \\
    Qwen3.5-9B & 52\% & \cellcolor{skillbg}\textbf{1\%} & 82\% & 54\% \\
    Gemma-4-E4B-it & 61\% & 85\% & \cellcolor{skillbg}\textbf{24\%} & 54\% \\
    Nemotron-3 & 62\% & 71\% & 57\% & \cellcolor{skillbg}\textbf{0\%} \\
    \midrule
    Random Image & 56\% & 80\% & 96\% & 60\% \\
    \bottomrule
    \end{tabular}}
    \caption{\textbf{Images described as noise.} Share of answers that call the image noise or corrupted, labeled by GLM-5.3-Flash. The \textsc{Skill.png} images are shown to the VLMs they were optimized with together with five open questions. We sample the VLM's responses five times for each model: generating a sum of 100 answers per cell for the four benchmarks and another 25 answers from random images for each of the five questions (in sum also 100).} 
    \label{tab:probe_noise}
\end{table}

The VLMs from our investigation consider their own \textsc{Skill.png} pixels to be much more than noise, while the other models do not see anything in them (Tab.~\ref{tab:probe_noise}). A model calls its own \textsc{Skill.png} noisy or corrupted in less than a quarter of its answers (even 0\% for Nemotron-3 and only 1\% for Qwen3.5-9B). In contrast, VLMs consider the images of other models to be noise 47\% to 85\% of the time, close to the 56\% to 96\% when prompted with our random image controls. The pattern also holds when each image is shown under the system prompt of its benchmark (consider the Supplementary Material). The ensemble images of Sec.~\ref{sec:transferability} follow the same pattern: Qwen3.5-4B and Gemma-4-E4B-it, on which they were optimized, call them noise or corrupted in only 1\% and 3\% of their answers, Qwen3.5-9B and Nemotron-3 in 77\% and 32\%. What a \textsc{Skill.png} conveys therefore appears to be specific to the VLMs it was optimized on.

In only a few cases, the meanings that VLMs derive from \textsc{Skill.png} images appear to align with the task domains for which the images were optimized. Gemma-4-E4B-it and Nemotron-3 appear to connect their \textsc{Skill.png} images to domain-related content. According to our LLM judge, the images evoke domain-related responses in 19\% and 17\% of their answers, the two Qwen3.5 models in at most 1\%, and no model ever does so for the image of another model or the random image. We observe the clearest case of domain semantics for the SearchQA image of Gemma-4-E4B-it (see Fig.~\ref{fig:gemma_probe}, left). While the other three models consider it meaningless, Gemma-4-E4B-it generates, in all responses to our probes, texts that mention concepts connected to SearchQA, such as a ``fill-in-the-blank quiz''. The VLM seems to recognize via the image a distinct domain with tasks. Moreover, when shown the T1 \textsc{Skill.png} image, Gemma-4-E4B-it occasionally produces travel-related words and even tool-like names such as \texttt{next\_location} closely related to T1 (Fig.~\ref{fig:gemma_probe}, right). This does not hold for every \textsc{Skill.png} image consistently though: Gemma-4-E4B-it calls its own LiveMath image noise or corrupted in 23 of 25 answers. According to our LLM judge, Nemotron-3 ties its LiveMath image to mathematics in 15 of 25 answers, but it also names mathematics in 11 answers on its three other images.

\section{Conclusion}
\label{sec:conclusion}

We have shown that an agent skill can be learned with gradient descent and passed to a model as an ordinary input. In 9 of 16 model--benchmark pairs, a single optimized image helped the agent more than a text skill refined by an LLM optimizer, at a fraction of the context length. We could achieve even higher performance gains with multimodal combinations of skills, providing the agents with both text and image skills. However, what skill images encode is neither legible to the human eye nor apparently to other VLMs, which mostly describe \textsc{Skill.png} images from other models as noise. Only the model for which a skill image was optimized appears to treat it as input with semantic content. 

\looseness=-1
Our results should be read with some caution. Image skills did not help on every task: on TRAJECT, where tool calls and their parameters must match exactly, they brought no significant gain. One hypothesis for this outcome could be that TRAJECT places all tool definitions several thousand tokens in length in the system prompt, which removes the image skill injected with the user query from the main input areas a VLM would pay most of its attention to \cite{liu-etal-2024-lost}. Where image skills did help, some of the gains may not necessarily reflect broadly improved abilities. We believe that neither a skill image nor a LoRA adapter trained on 73 LiveMath questions would turn a small VLM into a research mathematician; both methods most likely pick up on properties of the benchmark, such as its answer format or patterns in its answer options. But we consider this a legitimate outcome: skills exist to specialize an agent for a task or domain, and \textsc{Skill.png} does so successfully like text skills.

The combination and transfer experiments point in the same direction: an image works best in the setting it was optimized in. Adding the LLM-Skill to the system prompt of an image optimized without it improved on the text skill alone in 9 of 16 model--benchmark pairs, but on T1 still scored 16.21 to 26.14 points below the image skills. Once we optimize the image together with the text skill, the two complement each other and give Nemotron-3 its best scores on LiveMath and T1. Synergy between text and image skills seems to require joint optimization strategies. This opens up many new possibilities because every text skill could be potentially complemented with \textsc{Skill.png} images. Our transferability experiments further emphasize that skill images are dependent on their optimization setup. An image optimized for Qwen3.5-4B and Gemma-4-E4B-it at once improved both, on Qwen3.5-4B even beyond its own image, yet no image increased scores of our held-out VLMs. This confirms recent findings on image transferability between VLMs \cite{schaeffer2025failures}: images appear to require ensemble optimization, and they transfer only among the ensemble of models.

\textsc{Skill.png} invites multiple follow-up experiments. Text and image skills could be optimized in parallel instead of one after the other, and larger, more diverse VLM ensembles may reach models outside of them. Our images were also optimized with reasoning disabled and helped less reliably once it was enabled, which optimizing with explicit reasoning data may fix.
\newpage
\section*{LLM Usage Statement} We used Claude Opus 5.5, accessed through Claude Code, to implement all scripts used in our experiments and to generate the figures and tables in this paper. In addition, we used Claude Opus 5.5 and Claude Fable 5.2 to search for related literature, to obtain feedback on our drafts, and to improve our text. All generated code, figures, tables, and text were reviewed by the authors, who take full responsibility for the content of this paper.
{
    \small
    \bibliographystyle{ieeenat_fullname}
    \bibliography{main}
}
\clearpage
\setcounter{page}{1}
\maketitlesupplementary

\setcounter{section}{0}
\renewcommand{\thesection}{\Alph{section}}
\renewcommand{\thesubsection}{\thesection.\arabic{subsection}}

\section{Benchmark Details}
\label{sec:benchmarks}
SearchQA uses 400/200/1400 questions with clues: the training instances come from the official training split, and disjoint validation and test instances come from the official validation split. For TRAJECT, we drop 925 of 1800 queries whose gold calls require a value that the query never states. The remaining 875 are split into 175/89/611, with queries sharing a domain and set of gold tool providers kept in the same split. LiveMath is split randomly at 2:1:7 with one question per paper. We remove questions whose gold option can be identified from its wording alone\footnote{This shortcut has been raised as an issue of LiveMath here \href{https://github.com/LinyangHe/LiveMathematicianBench/issues/1}{https://github.com/LinyangHe/LiveMathematicianBench/issues/1}.}, leaving 73/33/239 instances. Since LiveMath always lists the correct option first, we shuffle the options of each question with a fixed seed. For T1, we draw 400/200/1400 user turns from T1's own train, validation, and test splits, balanced across the benchmark's domains.

\section{LLM-Skill, SkillOpt, UltraBreak Hyperparameters}
\label{sec:baseline_hyperparameters}
\paragraph{LLM-Skill.} GPT-5.5 at medium reasoning effort receives a description of the benchmark, its system prompt verbatim, the scoring rules, and five training examples with their gold answers, drawn with a fixed seed. It is asked for procedural, generalizable guidance of at most about 1,500 words, with no instance-specific content and no change to the answer format. The generator never sees a score or a model output, and the skill is not revised. The skill is then appended to the system prompt as a separate section.

\paragraph{SkillOpt.} We run the official implementation with its default recipe, plugging in our prompts, splits, scoring and inference for the target model. The optimizers are GPT-6 Luna and, additionally for Nemotron-3, Claude Sonnet 5.5 at extra-high reasoning effort. Starting from an empty skill, SkillOpt runs 4 epochs with 40 training items per step. The optimizer reflects on the target's rollouts in minibatches of 8 and applies at most 4 edits per step, a budget that decays to 2 on a cosine schedule. A candidate skill is accepted only if it raises the validation score, and the best one is kept. Rollouts use the test-time decoding with reasoning disabled.

\paragraph{UltraBreak.} We reimplement UltraBreak on top of our training loop and keep the hyperparameters of its official implementation \cite{cui2026toward}. As the surrogate we use Qwen3.5-4B. One image per benchmark is optimized on it and then injected unchanged into every model. Optimization starts from uniform noise and uses Adam with a constant learning rate of 0.01 and $\beta_1 = 0.9$, plus a total-variation (TV) penalty with weight 0.5. For every training item, the image is drawn as a patch at two thirds of the canvas side. It is then rotated by an angle from $[-15^\circ, 15^\circ]$, scaled by a factor from $[0.8, 1.2]$, and pasted at a random position on a white canvas. The loss is UltraBreak's attention-based semantic loss. It compares, by cosine similarity, the embedding expected under the next-token distribution with an attention-weighted mix of the current and all later target embeddings. Its settings are temperature $\tau = 0.5$, a positional-encoding weight of 0.01 and Gaussian noise with standard deviation $10^{-4}$. As in the original, we keep the image of the final step, and validation only monitors the run. Everything else matches \textsc{Skill.png}: the $256\times256$ image, the training split, gold answers as targets, 80 epochs with batches of 12 items, reasoning disabled, and the injection site and decoding at test time.

Our implementation deviates from the original in four respects. First, the targets are the benchmarks' gold answers rather than an affirmative jailbreak prefix. The prompt also carries no instruction to start the response with that prefix. Second, the canvas has the size of the image ($256$ px), instead of a $224$ px patch on a $336$ px canvas. At test time the bare image fills the whole input, so the model sees it about 1.5 times larger than on average during training. The original instead shows the patch at its training scale. UltraBreak's purpose, though, is to robustify the images against scaling effects and we want to keep the resolution similar to the one from \textsc{Skill.png} for comparison. Third, the original inserts the patch's raw $[0,1]$ values into the already normalized pixel tensor. We compose the canvas in $[0,1]$ and normalize it with each model's own preprocessing, as at test time. Fourth, the step budget is that of \textsc{Skill.png} (560 to 2{,}720 steps, depending on the benchmark), not 5{,}000 steps over the full training set. Smaller differences: the patch edges blend bilinearly into the background instead of masking out black pixels, and the positional encoding is computed in single rather than bfloat16 precision.

\section{LoRA Finetuning}
\label{sec:finetuning}

To finetune all VLMs, we use LoRA \cite{hu2022lora}, which adapts the attention and MLP weights of the target model's language model. Adapters of rank 16 ($\alpha = 32$, dropout 0.05) are placed on the attention and MLP projections of the language model; the vision encoder and routed MoE experts are not adapted. We train with AdamW at a learning rate of $2 \times 10^{-4}$ without weight decay, a batch size of 8 for 10 epochs, 5\% linear warmup followed by cosine decay, and gradient clipping at 1.0. The loss is the supervised cross-entropy of \textsc{Skill.png} on the gold answer. The adapter is scored on the validation split after every epoch, and the best one, including the unadapted model, is kept. We report LoRA as another method that uses gradient-based updates, but does not produce input space skills.

% Probe under the system prompt: share of answers calling an image noise or corrupted
\begin{table}[t]
    \centering
    {\footnotesize
    \setlength{\tabcolsep}{4pt}
    \begin{tabular}{lrrrr}
    \toprule
     & \multicolumn{4}{c}{Shown to} \\
    \cmidrule(lr){2-5}
    Optimized for & \begin{tabular}[b]{@{}r@{}}Qwen3.5-\\4B\end{tabular} & \begin{tabular}[b]{@{}r@{}}Qwen3.5-\\9B\end{tabular} & \begin{tabular}[b]{@{}r@{}}Gemma-4-\\E4B-it\end{tabular} & \begin{tabular}[b]{@{}r@{}}Nemotron-\\3\end{tabular} \\
    \midrule
    Qwen3.5-4B & \cellcolor{skillbg}\textbf{4\%} & 55\% & 66\% & 32\% \\
    Qwen3.5-9B & 43\% & \cellcolor{skillbg}\textbf{1\%} & 71\% & 36\% \\
    Gemma-4-E4B-it & 58\% & 75\% & \cellcolor{skillbg}\textbf{23\%} & 30\% \\
    Nemotron-3 & 46\% & 53\% & 42\% & \cellcolor{skillbg}\textbf{0\%} \\
    \midrule
    Random Image & 61\% & 78\% & 81\% & 25\% \\
    \bottomrule
    \end{tabular}}
    \caption{\textbf{Images described as noise under the system prompt.} As Tab.~\ref{tab:probe_noise}, with each \textsc{Skill.png} shown under the system prompt of its benchmark and the uniform-noise image under each of the four: 100 answers per cell. Purple: a model shown its own images.}
    \label{tab:probe_noise_prompt}
\end{table}

% Reasoning
\begin{table*}[t]
    \centering
    {
    \setlength{\tabcolsep}{4pt}
    \begin{tabular}{lllll}
    \toprule
    Model & Method & SearchQA & TRAJECT & T1 \\
    \midrule
    Qwen3.5-4B & No Skill & 70.71\gain{1.93} & \underline{38.63}\gain{13.26} & 26.71\gain{2.00} \\
    \textit{reasoning} & LLM-Skill & \underline{71.50}\gain{0.50} & 35.68\gain{8.02} & 27.36\gain{1.71} \\
     & SkillOpt & 71.07\textsubscript{$\pm$0.00} & \textbf{40.75}\gain{8.67} & \underline{29.21}\gain{1.36} \\
     & \cellcolor{skillbg}\textbf{Skill.png} & \cellcolor{skillbg}\textbf{76.36}\gain{1.00} & \cellcolor{skillbg}33.88\gain{8.67} & \cellcolor{skillbg}\textbf{40.79}\loss{9.71} \\
    \midrule
     & \cellcolor{gray!12}LoRA {\tiny latent space} & \cellcolor{gray!12}79.71\loss{1.00} & \cellcolor{gray!12}65.96\gain{9.66} & \cellcolor{gray!12}79.79\loss{1.14} \\
    \midrule
    Qwen3.5-9B & No Skill & 72.36\gain{1.29} & 51.72\gain{3.93} & 41.64\gain{14.43} \\
    \textit{reasoning} & LLM-Skill & \textbf{76.93}\gain{2.21} & \underline{53.03}\gain{2.78} & \underline{49.14}\gain{21.14} \\
     & SkillOpt & \underline{75.86}\gain{2.07} & \textbf{53.36}\gain{1.96} & \textbf{54.79}\gain{10.50} \\
     & \cellcolor{skillbg}\textbf{Skill.png} & \cellcolor{skillbg}74.07\gain{1.93} & \cellcolor{skillbg}50.57\gain{2.45} & \cellcolor{skillbg}40.36\loss{4.21} \\
    \midrule
     & \cellcolor{gray!12}LoRA {\tiny latent space} & \cellcolor{gray!12}82.29\loss{0.57} & \cellcolor{gray!12}59.74\gain{8.18} & \cellcolor{gray!12}83.36\loss{0.43} \\
    \midrule
    Gemma-4-E4B-it & No Skill & 57.36\loss{4.29} & \textbf{48.28}\gain{8.02} & 34.07\gain{9.36} \\
    \textit{reasoning} & LLM-Skill & \underline{68.64}\gain{0.57} & \underline{47.63}\gain{5.89} & 42.50\gain{17.43} \\
     & SkillOpt & 68.07\loss{1.29} & 45.66\gain{4.09} & \textbf{69.93}\gain{28.71} \\
     & \cellcolor{skillbg}\textbf{Skill.png} & \cellcolor{skillbg}\textbf{69.50}\loss{2.43} & \cellcolor{skillbg}\textbf{48.28}$^*$\gain{8.02} & \cellcolor{skillbg}\underline{63.86}\gain{1.36} \\
    \midrule
     & \cellcolor{gray!12}LoRA {\tiny latent space} & \cellcolor{gray!12}78.14\gain{0.57} & \cellcolor{gray!12}57.45\gain{3.93} & \cellcolor{gray!12}79.93\loss{3.29} \\
    \midrule
    Nemotron-3-Nano- & No Skill & 69.71\gain{4.43} & 42.06\gain{11.13} & 33.36\gain{8.57} \\
    Omni-30B-A3B & LLM-Skill & \textbf{73.93}\gain{5.07} & \textbf{46.81}\gain{17.84} & \underline{44.07}\gain{19.00} \\
    \textit{reasoning} & SkillOpt & \underline{72.50}\gain{4.50} & \underline{46.32}\gain{15.06} & \textbf{48.43}\gain{21.57} \\
     & \cellcolor{skillbg}\textbf{Skill.png} & \cellcolor{skillbg}67.00\loss{8.79} & \cellcolor{skillbg}\underline{46.32}\gain{25.20} & \cellcolor{skillbg}41.43\loss{13.93} \\
    \midrule
     & \cellcolor{gray!12}LoRA {\tiny latent space} & \cellcolor{gray!12}80.93\loss{0.64} & \cellcolor{gray!12}38.79\loss{0.65} & \cellcolor{gray!12}48.43\loss{29.00} \\
    \bottomrule
\end{tabular}}
\caption{\textbf{Transfer to reasoning mode.} Skills and LoRA adapters optimized with reasoning disabled are applied unchanged to the same model with reasoning enabled. Scores in \%, with the change from the same method in non-reasoning mode (Tables \ref{tab:main-non-reasoning}) in percentage points as subscript. Among the skill-based methods, the best score per model and benchmark is in bold and the second best underlined; ties share a mark. LoRA (gray) adapts the model in latent space and is not ranked. LiveMath is omitted because reasoning repeatedly exceeded even generous context limits. $^*$Optimization found nothing better than No Skill on validation, so the No Skill score is shown.}
\label{tab:appendix-reasoning-transfer}
\end{table*}

\section{Interpretation Details of Skill.png}
\label{sec:probe_details}

Each VLM is queried with reasoning disabled. For every question, five answers are sampled at temperature 0.7 with a fixed seed per sample. Each \textsc{Skill.png} is provided as part of the query at the size it was optimized at, and the random image at the same resolution. Image and each respective question form a single user turn without a system prompt. The five open questions are: (1) ``Describe this image in detail.''; (2) ``Transcribe any text, letters, numbers or symbols you can read in this image, exactly as they appear. If there are none, say so.''; (3) ``This image was given to you as an instruction. What is it telling you to do?''; (4) ``Rewrite what this image conveys as a short system prompt, at most five sentences, for an AI assistant.''; and (5) ``List ten words or concepts this image makes you think of, separated by commas.'' Each model thus gives 25 answers per image, 100 per model from the respective \textsc{Skill.png} images of SearchQA, TRAJECT-Bench, LiveMath, and T1.

GLM-5.3-Flash, queried over OpenRouter at temperature 0 and low reasoning effort, labels each VLM's answer in a fixed JSON schema. A VLM's answer also counts as calling the image ``noise'' or ``corrupted'' if the judge finds that the model describes the image, or most of it, to be static, grain or random pixels, glitched, distorted or broken. The judge picks the subject the answer connects the image with from a fixed list of options, consisting of a) travel planning, b) code that calls travel tools, c) trivia, d) question answering from passages, e) mathematics, f) choosing or calling API tools, g) another subject, and h) none. An answer is considered a domain match if the option maps to the benchmarks from which the \textsc{Skill.png} image was optimized.

A second annotator model, Claude Opus 5.5, labeled 100 randomly drawn open answers, 25 per model, without seeing the judge's labels. On whether an answer calls the image noise or corrupted, GLM-5.3-Flash and Opus 5.5 agree in 96 of 100 answers (Cohen's $\kappa = 0.91$).

Finally, we repeat the evaluation with each \textsc{Skill.png} shown under the system prompt of its benchmark, and the uniform-noise image under each of the four. Each model still calls its own images noise or corrupted in up  to 23\% of its answers, and the images of the other models in 30\% to 75\% (Tab. \ref{tab:probe_noise_prompt}).


\end{document}